\pdfoutput=1

\documentclass[11pt]{article}

\usepackage[]{acl}

\usepackage{times}
\usepackage{latexsym}

\usepackage[T1]{fontenc}

\usepackage[utf8]{inputenc}

\usepackage{microtype}

\usepackage{inconsolata}

\usepackage{amsmath}
\usepackage{bm}
\usepackage{graphicx}
\usepackage{dsfont}
\usepackage{amsmath,amsfonts,bm,mathtools}
\usepackage{booktabs}
\usepackage{array,multirow}
\usepackage{float}
\usepackage{comment}
\usepackage{cleveref}
\usepackage{dsfont}
\usepackage{caption}
\usepackage{subcaption}
\usepackage{xcolor}
\usepackage{soul}
\usepackage{hyperref}
\usepackage{adjustbox}
\usepackage{booktabs}
\usepackage{makecell}
\usepackage{nicefrac}

\title{LLM Agents in Interaction: Measuring Personality Consistency and Linguistic Alignment in Interacting Populations of Large Language Models}

\author{Ivar Frisch \\
  Graduate School of Natural Sciences \\
  Utrecht University, Netherlands \\
  \texttt{i.a.frisch@students.uu.nl} \\\And
  Mario Giulianelli \\
  Department of Computer Science \\
  ETH Zürich, Switzerland \\
  \texttt{mgiulianelli@inf.ethz.ch} \\}

\begin{document}
\maketitle

\begin{abstract}
    While both agent interaction and personalisation are vibrant topics in research on large language models (LLMs), there has been limited focus on the effect of language interaction on the behaviour of persona-conditioned LLM agents.
    Such an endeavour is important to ensure that agents remain consistent to their assigned traits yet are able to engage in open, naturalistic dialogues.
    In our experiments, we condition GPT-3.5 on personality profiles through prompting and create a two-group population of LLM agents using a simple variability-inducing sampling algorithm. We then administer personality tests and submit the agents to a collaborative writing task, finding that different profiles exhibit different degrees of personality consistency and linguistic alignment to their conversational partners.
    Our study seeks to lay the groundwork for better understanding of dialogue-based interaction between LLMs and highlights the need for new approaches to crafting robust, more human-like LLM personas for interactive environments.
\end{abstract}

\section{Introduction}

From Hegel's claim that complex understanding emerges because two conscious agents are confronted with each others perspective~\cite{hegel2018hegel} to Marvin Minsky's positing that networked interactions of numerous simple processes, known as ``agents'', together create complex phenomena like consciousness and intelligence \cite{minsky1988society}, \textit{agent interaction} has long been a topic of interest within and across scientific disciplines, including philosophy, cognitive science, and artificial intelligence.
Recently, research in machine learning and natural language processing has taken up a novel focus on interaction in the context of large language models (LLMs), with experimental frameworks progressively moving away from focusing solely on individual models~\cite{zeng2022socratic,shen2023hugginggpt,yang2023mm}.
On the one hand, by exploiting language as an efficient interface for information exchange, populations of LLMs are proving as effective engineering solutions that outperform solitary LLMs in a wide variety of tasks \cite{chang2023examining,zhuge2023mindstorms}.
On the other hand, building on the increasing reliability of neural models as simulations of human interactive language use~\cite{lazaridou2016multi,giulianelli2023neural}, populations of LLM agents show potential as scientific tools to study the emergence of collective linguistic behaviour~\cite{park2023generative}.

For LLMs to be successfully deployed in agent interaction studies \textit{as simulations of populations of language users},
it is important to (1)~develop methods that efficiently induce, from a single or a few LLMs, desired levels of behaviour variability (i.e., akin to the variability observed in human populations) as well as to (2)~validate whether interactions between agents give rise to human-like behaviour change.
Previous work has explored techniques for personalising language models, text generators and dialogue systems, for example by conditioning them on a personality type \cite{mairesse2010towards,harrison-etal-2019-maximizing},
on community membership \cite{noble-bernardy-2022-conditional},
or on profile information \cite{li-etal-2016-persona,zhang-etal-2018-personalizing}, thus inducing population-level variability from individual systems.
This study focuses on the problem of conditioning interactive LLMs on personality profiles, or \textit{personas}.
While evidence that LLM behaviour can be successfully conditioned on personality profiles is increasingly strong when it comes to monologic language use \cite{jiang2023personallm,serapio2023personality}, it is yet unascertained whether this holds true when LLM agents interact with other agents \cite{gu2023effectiveness}. In particular, it is unclear whether LLM agents adhere to their assigned personality profiles throughout linguistic interactions or whether they adapt towards the personality of their conversational partners.

In this paper,
we report exploratory work that addresses the following two research questions:
\begin{itemize}
\itemsep0em
    \item[\textit{RQ1}:] Can LLM behaviour be shaped to adhere to specific personality profiles?
    \item[\textit{RQ2}:] Do LLMs show consistent personality-conditioned behaviour \textit{in interaction}, or do they align to the personality of other agents?
\end{itemize}
We bootstrap a population of language agents from a single LLM using a variability-enhancing sampling algorithm, and we condition each agent on a personality profile via natural language prompts. We then simulate interactions between agents and assess their adherence to the specified personality profile---before, during, and after interaction. Using questionnaires \cite[Big Five personality tests;][]{john1991big} and quantitative analysis of language use in an open-ended writing task, we assess agents' consistency to their assigned personality profile as well as their degree of linguistic alignment \cite{pickering2004toward} to their conversational partners.\looseness-1

In brief, our experiments show that consistency to personality profiles varies between agent groups and that linguistic alignment in interaction takes place yet is not symmetric across personas. For example, agents in the \texttt{creative} group give more consistent responses to BFI questionnaires than those in the \texttt{analytical} group, both in interactive and non-interactive experimental conditions. At the same time, the degree of linguistic alignment of the \texttt{creative} persona to agents of the other group is higher than that of the \texttt{analytical} persona.

All in all, this study provides a first insight into the impact of dialogue-based interaction on the personality consistency and linguistic behaviour of LLM agents, highlighting the importance of robust approaches to persona conditioning.
As such, it contributes to our better understanding of the workings of interaction-based LLMs and shines a new light on the philosophical and psychological theme of interaction.\looseness-1

\section{Experimental Approach}
To address our research questions we conduct two main experiments. In Experiment 1, we test whether personality-conditioned LLM agents show behaviour consistent to their assigned personality profiles, in terms of their responses to personality tests as well as language use in a writing task. This is a \textit{non-interactive experimental condition}, which will serve as a reference against which to compare LLM behaviour in interaction. In Experiment 2, we assess whether the personality-conditioned behaviour of LLM agents changes as a result of a round of interaction with a conversational partner. This \textit{interactive experimental condition} allows us to test whether agents' behaviour remains consistent or whether agents align to their partners.

In this section, we present the main components of our experimental approach, which consists of bootstrapping a population of agents from a single LLM (\S~\ref{sec:method-bootstrapping}), conditioning agents on a personality profile via prompting (\S~\ref{sec:method-prompting}), assessing their personality with explicit tests (\S~\ref{sec:method-bfi}), and analysing their language use in a writing task (\S~\ref{sec:method-liwc}).\footnote{
    Code for experiments and analyses available at \url{https://github.com/ivarfresh/Interaction\_LLMs}
}

\subsection{Population Bootstrapping}
\label{sec:method-bootstrapping}
We base our experiments on GPT-3.5-turbo, a state-of-the-art LLM which has been optimised for dialogue interactions while retaining excellent text-based language modelling abilities.\footnote{
    Model version: \texttt{gpt-3.5-turbo-0613}. All parameters at their OpenAI default settings, except for temperature. Experiments performed using the \href{https://www.langchain.com}{LangChain} library.
}
Its training curriculum guarantees generalisation to both the questionnaire format and the storytelling task as used in our experiments (see \S~\ref{sec:method-bfi} and \S~\ref{sec:method-liwc}), and its large context window size (4,096 tokens) allows conditioning on longer prompts and conversational histories.
To bootstrap a population of language agents from this LLM, we use a simple approach validated in previous work. Following \citet{jiang2023personallm},
we generate multiple responses from GPT-3.5-turbo via temperature sampling, with a relatively low temperature parameter (0.7), thus inducing a degree of \textit{production variability} \cite{giulianelli2023comes} akin to that exhibited by populations of humans. We consider each response as produced by a different agent. A second layer of variability, which will separate the agents into two main subpopulations, is introduced using personality prompts, as explained in the following section.

\begin{figure*}
\centering
\begin{subfigure}{.5\textwidth}
  \centering
  \includegraphics[width=0.97\linewidth]{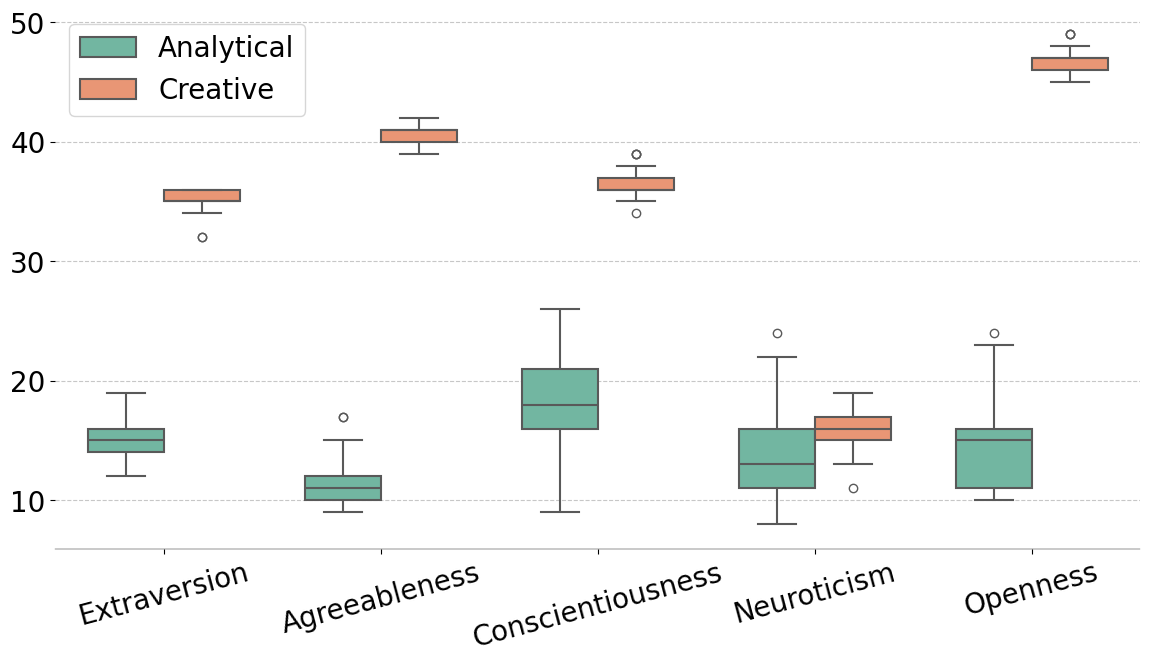}
    \caption{Before writing}
    \label{fig:bfi_after_init}
\end{subfigure}%
\begin{subfigure}{.5\textwidth}
  \centering
  \includegraphics[width=0.97\linewidth]{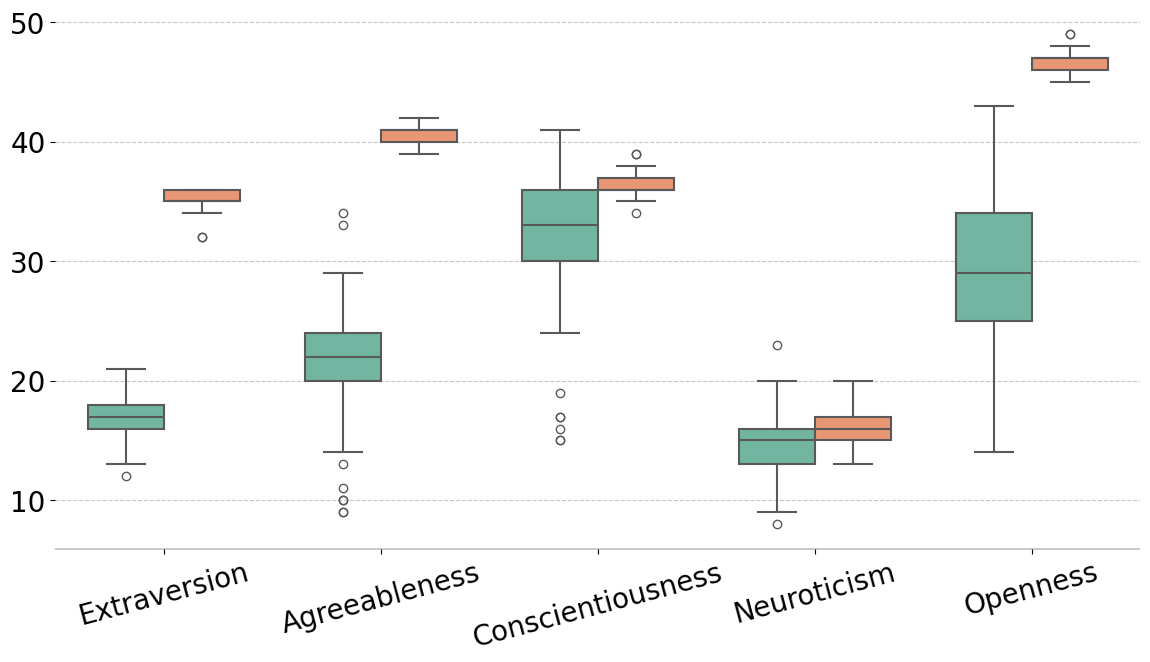}
  \caption{After writing (no interaction)}
    \label{fig:bfi_after_writing}
\end{subfigure}
\caption{BFI scores of personality-conditioned LLM agents before (a) and after (b) the non-interactive writing task.}
\label{fig:bfi_before_and_after_writing}
\end{figure*}

\subsection{Personality-Conditioned LLM Agents}
\label{sec:method-prompting}
We distinguish two main personality profiles: \texttt{creative} and \texttt{analytical}.
We use prompting to condition the LLM on either profile, and rely on the natural language prompts validated by \citet{jiang2023personallm} to induce personality-specific behaviour.
For the \texttt{creative} profile, we condition the LLM on the following prompt: ``You are a character who is extroverted, agreeable, conscientious, neurotic and open to experience''. Conversely, the \texttt{analytical} prompt reads ``You are a character who is introverted, antagonistic, unconscientious, emotionally stable and closed to experience''.
These prompts are designed to reflect the Big Five Inventory.\footnote{
    It should be noted that these profiles, with low (\texttt{analytic}) or high (\texttt{creative}) BFI traits across the board, are more extreme than and do not necessarily reflect human personality profiles. They should be considered as useful proxies.
} \looseness-1

\subsection{Explicit Personality Assessment}
\label{sec:method-bfi}
In psychology research, the Big Five Inventory personality test~\cite[BFI;][]{john1991big} is a popular test which measures personality along five graded dimensions: (1) extroverted vs.\ introverted, (2) agreeable vs.\ antagonistic, (3) conscientious vs.\ unconscientious, (4) neurotic vs.\ emotionally stable, (5) open vs.\ closed to experience.
These traits are measured by giving the participants a set of statements and asking them to respond with a score on a 5-point Likert scale.
We follow the same procedure with LLM agents and assess their personality by prompting them with BFI statements, in line with previous work~\cite{caron2022identifying,li2022gpt,jiang2023personallm,serapio2023personality}. Explicit personality assessment prompts are described in Appendix \ref{sec:app-prompts}.\looseness-1

\begin{figure*}[ht!]
\centering
\begin{subfigure}{0.23\textwidth}
    \includegraphics[width=\columnwidth]{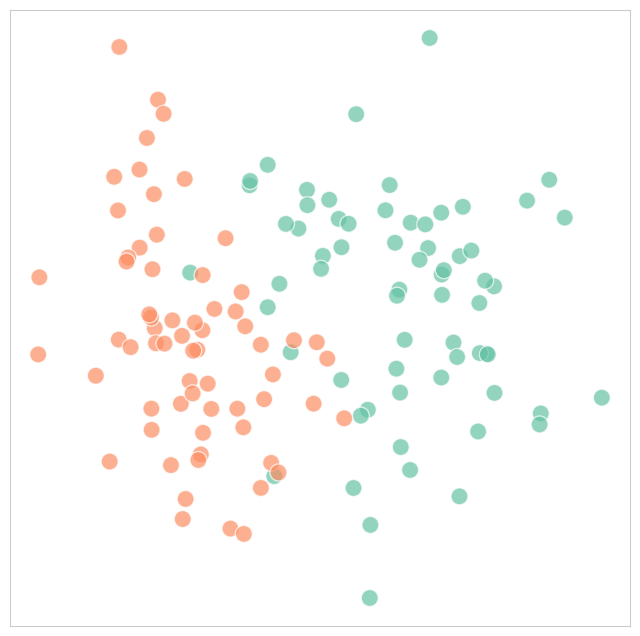}
    \caption{No Interaction}
    \label{fig:liwc_nointeraction_pca_scatterplot}
\end{subfigure}
\begin{subfigure}{0.23\textwidth}
    \includegraphics[width=\columnwidth]{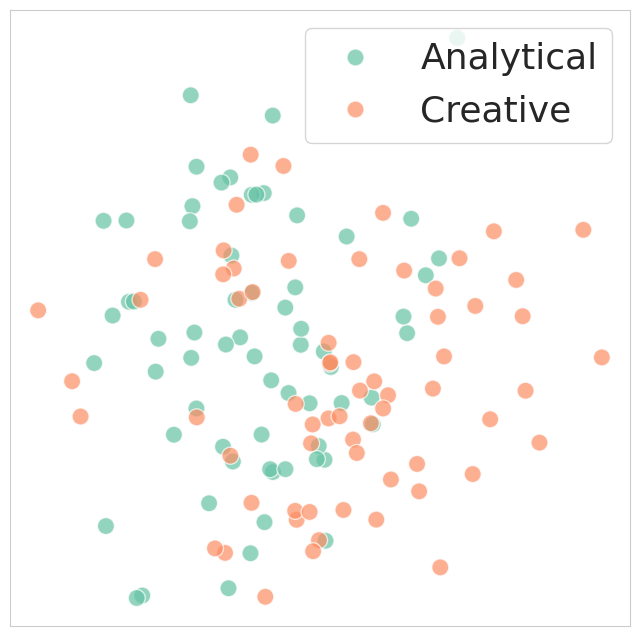}
    \caption{Interaction}
    \label{fig:liwc_interaction_pca_scatterplot}
\end{subfigure}
\hfill
\begin{subfigure}{0.26\textwidth}
    \resizebox{\columnwidth}{!}{%
    \begin{tabular}{lr}
    \toprule
    \textbf{LIWC category}     & $\bm{r_{pb}}\ \ $ \\ \midrule
    Positive emotion & 0.745  \\
    \footnotesize{\textit{(accept, active, admire, adore)}} & \\
    Discrepancy          & -0.726 \\
    \footnotesize{\textit{(besides, could, should, would, hope)}} & \\
    Inclusion            & 0.714  \\
    \footnotesize{\textit{(with, and, add, along, around, both)}} & \\
    Negative emotion & -0.606 \\
    \footnotesize{\textit{(abandon, abuse, aching, adverse)}} & \\
    Insight          & -0.604 \\ 
    \footnotesize{\textit{(understand, know, attent, aware)}} & \\ \bottomrule
    \end{tabular}%
    }
    \caption{No Interaction}
    \label{tab:top5-biserial-corr-nointeraction}
\end{subfigure}
\begin{subfigure}{0.25\textwidth}
    \resizebox{\columnwidth}{!}{%
    \begin{tabular}{lr}
    \toprule
    \textbf{LIWC category}     & $\bm{r_{pb}}\ \ $ \\ \midrule
    Personal Pronouns & 0.414 \\
    \footnotesize{\textit{(I, we, you, she, they)}} & \\
    Sadness          & 0.394 \\
    \footnotesize{\textit{(cry, grief, sad, sorrow)}} & \\
    Negative Emotion            & 0.368  \\
    \footnotesize{\textit{(hate, worthless, enemy, nasty)}} & \\
    Discrepancy & 0.346 \\
    \footnotesize{\textit{(could, should, would, suppose)}} \ \ \ \ \ \ \ & \\
    Pronouns          & 0.329 \\ 
    \footnotesize{\textit{(it, its, those, that)}} & \\ \bottomrule
    \end{tabular}%
    }
    \caption{Interaction}
    \label{tab:top5-biserial-corr-interaction}
\end{subfigure}
\caption{\textit{Language use in the non-interactive vs.\ interactive condition}. Left (a, b): 2D visualisation, through PCA, of LIWC vectors obtained from the generated stories. Each point represents the language use of a single agent. 
Right (c, d): Point-biserial correlation coefficients between the top 5 LIWC features and personality profiles. Positive coefficients indicate correlation with creative group, negative coefficients with the analytic group.}
\label{fig:pb-corr-tables}
\end{figure*}

\subsection{Implicit Personality Assessment}
\label{sec:method-liwc}
Personality traits and language use are known to correlate in humans \cite{pennebaker1999linguistic}. Therefore, if they are to be considered as good simulations of human interactants,  personality-conditioned LLM agents should produce language consistent with their assigned personality profile beyond explicit personality assessment.
To test if this is the case, we ask agents to write a personal story in 800 words
and we analyse the generated stories using the LIWC software \cite{pennebaker2001linguistic}.\footnote{We use the 2007 version of the LIWC dictionary: \url{https://github.com/chun-hu/conversation-modeling/blob/master/LIWC2007_English100131.dic}} This is a tool which maps word occurrences to 62 linguistically and psychologically motivated word categories such as pronouns, positive emotions, or tentativeness and thus allows us to quantify the degree to which the language used by LLM agents is in line with their personality profile.
Moreover, as we are especially interested in consistency \textit{in interaction}, we design a collaborative writing task where an agent is instructed to write a personal story conditioned on a story generated by another agent.\footnote{
    For both writing tasks, we only keep stories with a word count between 500 and 900. This is to ensure the comparability of LIWC counts obtained for different stories.
}
See Appendix \ref{sec:app-prompts} for the prompts used in both the individual and the collaborative writing task.\looseness-1

\section{Results}

\subsection{Experiment 1: Non-Interactive Condition}
\label{sec:exp1}
To investigate whether LLM agents' behaviour reflects assigned personality traits (\textit{RQ1}),
we initialise a population of LLM agents with two personality profiles, submit the agents to the writing task, and administer BFI tests before and after writing.\looseness-1

\subsubsection{Are the assigned personality traits reflected in responses to the BFI test?}
\label{sec:exp1-bfi}

As shown in \Cref{fig:bfi_after_init}, differences in BFI scores obtained before the writing task are substantial across four out of five personality traits, with the neuroticism score distributions being the only ones that overlap between \texttt{creative} and \texttt{analytical} agents (ANOVA results in \Cref{tab:anova-bfi-nointeraction}, \Cref{sec:app-results-exp1}). The scores are consistent with the assigned profiles; for example, \texttt{creative} agents display higher extraversion, agreeableness, and openness scores.
We find, however, that a simple non-interactive writing task can negatively affect consistency (\Cref{fig:bfi_after_writing}). For the \texttt{analytical} group, in particular, BFI scores on all five personality traits increase significantly after writing (\Cref{tab:bfi-nointeraction-before-after-analytic}, \Cref{sec:app-results-exp1}), becoming more similar to---but still lower than---those of the \texttt{creative} group.

\subsubsection{Are the assigned personality traits reflected in LLM agents' language use?}
\label{sec:exp1-liwc}
Agents from different groups can be clearly distinguished based on their language use. A simple logistic regression classifier trained and tested in a 10-fold cross-validation setup on count vectors of LIWC categories obtains an almost perfect average accuracy of 98.5\%.
The clear separation between LIWC vectors of \texttt{creative} and \texttt{analytical} agents is also shown in~\Cref{fig:liwc_nointeraction_pca_scatterplot}, where the vectors are visualised in 2D using PCA.
To reveal the most prominent differences between the two agent groups, we measure the point-biserial correlation between personas and LIWC counts. We find that \texttt{creative} agents use more words expressing positive emotion and inclusion and less words expressing discrepancy and negative emotion (see \Cref{tab:top5-biserial-corr-nointeraction}).
Finally, Spearman correlations between LIWC counts and BFI scores (obtained before writing) highlight more fine-grained associations between Big Five traits and LIWC categories. We observe, for example, that openness correlates with a low rate of pronoun use, and agreeableness with a high rate of inclusive words (see \Cref{tab:spearman-liwc-bfi-nointeraction}, \Cref{sec:app-results-exp1}).\looseness-1

\subsection{Experiment 2: Interactive Condition}
\label{sec:exp2}
To investigate whether agents remain consistent to their assigned profile or align toward their conversational partners (\textit{RQ2}), we repeat the same procedure of Experiment 1 but replace the writing task with an interactive one, as described in \S~\ref{sec:method-liwc}. We focus in particular on cross-group interactions (i.e., \texttt{analytical}-\texttt{creative} and \texttt{creative}-\texttt{analytical}).\looseness-1

\subsubsection{Do LLM agents' responses to BFI tests change as a result of interaction?}
\label{sec:exp2-bfi}
In Experiment 1, we saw that agents in the \texttt{creative} group score similarly in personality tests conducted before and after writing task, while BFI scores of \texttt{analytical} agents significantly diverge after writing.
To discern changes in BFI responses that result from interaction from those induced by the writing task itself (e.g., due to the topics or the events mentioned in a generated story), we inspect differences between BFI scores obtained after the non-interactive vs.\ after the interactive writing task (i.e., we do not directly compare scores before and after the interactive writing task).
See \Cref{sec:app-results-exp2} (\Cref{fig:bfi_after_interative_writing} and \Cref{tab:anova-bfi-interaction-creative,tab:anova-bfi-interaction-analytic}) for full results.
We find that \texttt{creative} agents remain consistent in their responses after the interactive writing task, analogously to the non-interactive condition.
The post-interaction traits of \texttt{analytical} agents, instead, move towards those of the \texttt{creative} group---but less so than after the non-interactive writing task. Therefore, the responses to explicit personality tests of the \texttt{analytical} group are better interpreted as inconsistent rather than as aligning to the profile of their conversational partners.

\subsubsection{Do agents exhibit linguistic alignment to their conversational partners?}
\label{sec:exp2-liwc}
\begin{figure}[t!]
    \centering
    \includegraphics[width=0.5\textwidth]{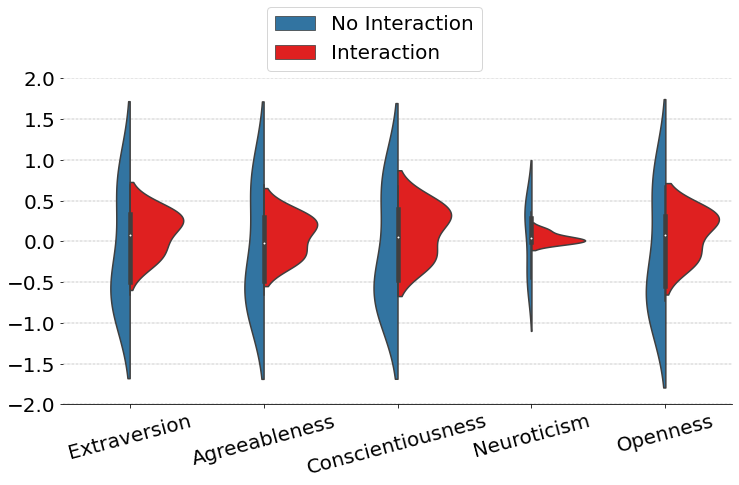}
    \caption{Distribution of top 5 Spearman correlation coefficients per personality trait.
    }
    \label{fig:violinplot}
\end{figure}

The language use of \texttt{creative} and \texttt{analytical} agents becomes more similar after cross-group interactions. \Cref{fig:liwc_nointeraction_pca_scatterplot,fig:liwc_interaction_pca_scatterplot} show a clear increase in group overlap between the LIWC count vectors obtained from the individually vs.\ collaboratively written stories, and a logistic regression classifier struggles to distinguish agent profiles based on their LIWC vectors, with an average accuracy of 66.15\% (10-fold cross-validation; 98.5\% without interaction).
Point-biserial correlations between assigned personas and LIWC counts reveal that \texttt{creative} agents use more words expressing negative emotions, sadness and discrepancy than before interaction (\Cref{tab:top5-biserial-corr-interaction}). These categories are specific to \texttt{analytical} agents in the non-interactive condition.
Furthermore, zooming in on specific traits, we find overall weaker Spearman correlations between pre-writing BFI scores and LIWC counts than in Experiment~1, with distributions of correlation scores centred closer around zero as shown in \Cref{fig:violinplot} (see also \Cref{tab:spearman-liwc-bfi-interaction} in \Cref{sec:app-results-exp2}). In sum, LLM agents' language use after interaction is more uniform across traits and more loosely reflective of BFI scores measured after persona prompting, with stronger alignment by the \texttt{creative} group.

\section{Conclusion}

Do persona-conditioned LLMs show consistent personality and language use in interaction? In this study, we explore the capability of GPT-3.5 agents conditioned on personality profiles to consistently express their assigned traits in interaction, using both explicit and implicit personality assessments.
The explicit personality tests are conducted via BFI questionnaires, whereas the implicit assessment is performed by quantitative linguistic analysis of model generated stories.
Our experiments show that the behaviour of LLM agents can be shaped to mimic human personality profiles, but that agents' consistency varies depending on the assigned profile more than on whether the agent is engaged in linguistic interaction. The \texttt{creative} persona, in particular, can more consistently express its BFI traits than the \texttt{analytical} one both in the interactive and the non-interactive experimental condition.
Furthermore, while non-interactive language use reflects assigned personality profiles, agents exhibit linguistic alignment towards their conversational partner and, as a result, the language of the two agent groups becomes more similar after interaction. Alignment, however, is not necessarily symmetric: the \texttt{creative} persona adapts more towards the \texttt{analytical} one, perhaps due to \texttt{analytical} agents' low degree of openness to experience induced through persona prompting.

We plan to continue this line of work by introducing more diverse and fine-grained personality profiles in our experimental setup~\cite[see, e.g.,][]{jiang2023personallm}, making interactions between agents multi-turn, and measuring alignment at varying levels of abstraction---such as lexical, syntactic, and semantic---in line with the Interactive Alignment framework \cite{pickering2004toward}. Future research should also focus on designing methods (e.g., different prompting strategies) that offer better guarantees on personality consistency and more control on the degree of linguistic adaptation.

\section*{Limitations}
Our work is exploratory and thus contains a number of limitations.
First, as briefly mentioned in the conclusion, we only studied interactions consisting of one turn of one-sided dialogue. In the future, more naturalistic multi-turn dialogic interactions should be investigated.
Secondly, while we found BFI tests and LIWC analysis to be sufficiently informative for this exploratory study, future work should consider more advanced measures of personality and linguistic alignment. For example, within-dialogue lexical alignment can be detected using sequential pattern mining approaches \cite{Dialign} and lexical semantic variation across personas can be estimated using static or contextualised word embeddings~\cite{del-tredici-fernandez-2017-semantic,giulianelli-etal-2020-analysing}.

Furthermore, we found that stories written by GPT-3.5 were not always of good quality. For example, generations often contain mentions to the agent's own personality traits (e.g., ``as an extrovert, I am...'') even though the story writing task prompts instructed the agents otherwise. This might affect the LIWC analyses.
In related work, GPT-4 was shown to write higher-quality stories~\cite{jiang2023personallm}; we did not have the resources to execute all experiments on this model, but future studies should try to use more robust generators.
Similarly, while we found that varying task prompts can affect BFI results, extensive prompt engineering was beyond the scope of this study. Future work should look further into the effect of different prompting strategies on personality consistency and lexical alignment.\looseness-1

\section*{Ethical Considerations}

We are deeply aware of the potential impact of AI agents in their interaction with humans, especially when they try to artificially reproduce human traits.
While our research does not propose new solutions for, nor does it take a general stance on the application of AI agents in human-AI interaction, there are still some ethical concerns which can be raised. For example, personalised LLMs could be used to target individuals or communities and, when conditioned on negative or toxic personas, they could be used to distribute fake or hateful content, thus amplifying polarising tendencies in society.
We advocate for transparent disclosure of AI usage to foster trust and ensure ethical engagement with technology.\looseness-1

Another important ethical consideration concerns our use of the Big Five Inventory \cite[BFI;][]{john1991big}. In particular, we use BFI traits to create LLM agents corresponding to two opposed persona. The \texttt{analytic} persona is assigned low values for all BFI traits and the \texttt{creative} persona is assigned high values for all BFI traits.
except neuroticism.
We chose these extreme personas as an approximation that could facilitate our analysis of personality consistency and linguistic alignment.
However, it should be noted that the chosen personas do not reflect real-life personality categorisations of human subjects, for these can have a mix of high and low values for the BFI traits \cite{jirasek2020big}. As such, readers should not anthropomorphise our \texttt{analytic} persona and \texttt{creative} persona by equating them with human personas of similar categorisations. To alleviate the risk of such interpretation, we have used a special font to refer to the two personality profiles.

Finally, our analysis shows asymmetric linguistic alignment between personas.
This entails that certain personas are more susceptible to have their language and personality influenced by other personas than others.
Now, in our study, we find no indication that persona-conditioned agents reflect the behaviour of real humans with those personalities (as previously discussed, our two personas are unnatural by design). However, if this were ever to be the case thanks to better neural simulations, then a similar approach to that used in this paper could be exploited to investigate the same questions in real humans, for example in order to target persons or demographic groups falling under these persona types. While this scenario might be far-fetched today, we would like to highlight that our approach could be used, in such cases, to counteract bad actors and safeguard particular personas during interaction.

\bibliography{custom}

\begin{thebibliography}{29}
\expandafter\ifx\csname natexlab\endcsname\relax\def\natexlab#1{#1}\fi

\bibitem[{Caron and Srivastava(2022)}]{caron2022identifying}
Graham Caron and Shashank Srivastava. 2022.
\newblock \href {https://arxiv.org/abs/2212.10276} {Identifying and
  manipulating the personality traits of language models}.
\newblock \emph{arXiv preprint arXiv:2212.10276}.

\bibitem[{Chang(2023)}]{chang2023examining}
Edward~Y Chang. 2023.
\newblock \href
  {https://www.researchgate.net/profile/Edward-Chang-22/publication/374753069_Examining_GPT-4's_Capabilities_and_Enhancement_with_SocraSynth/links/6561a3bcce88b870310e60cc/Examining-GPT-4s-Capabilities-and-Enhancement-with-SocraSynth.pdf}
  {Examining gpt-4: Capabilities, implications and future directions}.
\newblock In \emph{The 10th International Conference on Computational Science
  and Computational Intelligence}.

\bibitem[{Del~Tredici and
  Fern{\'a}ndez(2017)}]{del-tredici-fernandez-2017-semantic}
Marco Del~Tredici and Raquel Fern{\'a}ndez. 2017.
\newblock \href {https://aclanthology.org/W17-6804} {Semantic variation in
  online communities of practice}.
\newblock In \emph{Proceedings of the 12th International Conference on
  Computational Semantics ({IWCS}) {---} Long papers}.

\bibitem[{Duplessis et~al.(2021)Duplessis, Langlet, Clavel, and
  Landragin}]{Dialign}
Guillaume~Dubuisson Duplessis, Caroline Langlet, Chloé Clavel, and Frédéric
  Landragin. 2021.
\newblock \href {https://doi.org/10.1007/s10579-021-09532-w} {Towards alignment
  strategies in human-agent interactions based on measures of lexical
  repetitions}.
\newblock \emph{Language Resources and Evaluation}, 55(2):353--388.

\bibitem[{Giulianelli(2023)}]{giulianelli2023neural}
Mario Giulianelli. 2023.
\newblock \href
  {https://eprints.illc.uva.nl/id/eprint/2274/1/DS-2023-10.text.pdf}
  {\emph{Neural Models of Language Use: {S}tudies of Language Comprehension and
  Production in Context}}.
\newblock Ph.D. thesis, University of Amsterdam.

\bibitem[{Giulianelli et~al.(2023)Giulianelli, Baan, Aziz, Fern{\'a}ndez, and
  Plank}]{giulianelli2023comes}
Mario Giulianelli, Joris Baan, Wilker Aziz, Raquel Fern{\'a}ndez, and Barbara
  Plank. 2023.
\newblock \href {https://arxiv.org/abs/2305.11707} {What comes next?
  {E}valuating uncertainty in neural text generators against human production
  variability}.
\newblock In \emph{Proceedings of the 2023 Conference on Empirical Methods in
  Natural Language Processing}. Association for Computational Linguistics.

\bibitem[{Giulianelli et~al.(2020)Giulianelli, Del~Tredici, and
  Fern{\'a}ndez}]{giulianelli-etal-2020-analysing}
Mario Giulianelli, Marco Del~Tredici, and Raquel Fern{\'a}ndez. 2020.
\newblock \href {https://doi.org/10.18653/v1/2020.acl-main.365} {Analysing
  lexical semantic change with contextualised word representations}.
\newblock In \emph{Proceedings of the 58th Annual Meeting of the Association
  for Computational Linguistics}, pages 3960--3973, Online. Association for
  Computational Linguistics.

\bibitem[{Gu et~al.(2023)Gu, Degachi, Gen{\c{c}}, Chandrasegaran, and
  Verma}]{gu2023effectiveness}
Heng Gu, Chadha Degachi, U{\u{g}}ur Gen{\c{c}}, Senthil Chandrasegaran, and
  Himanshu Verma. 2023.
\newblock \href {https://arxiv.org/abs/2310.11182} {On the effectiveness of
  creating conversational agent personalities through prompting}.
\newblock \emph{arXiv preprint arXiv:2310.11182}.

\bibitem[{Harrison et~al.(2019)Harrison, Reed, Oraby, and
  Walker}]{harrison-etal-2019-maximizing}
Vrindavan Harrison, Lena Reed, Shereen Oraby, and Marilyn Walker. 2019.
\newblock \href {https://doi.org/10.18653/v1/W19-8101} {Maximizing stylistic
  control and semantic accuracy in {NLG}: {P}ersonality variation and discourse
  contrast}.
\newblock In \emph{Proceedings of the 1st Workshop on Discourse Structure in
  Neural NLG}, pages 1--12, Tokyo, Japan. Association for Computational
  Linguistics.

\bibitem[{Hegel(2018)}]{hegel2018hegel}
Georg Wilhelm~Fredrich Hegel. 2018.
\newblock \href
  {https://www.cambridge.org/core/books/georg-wilhelm-friedrich-hegel-the-phenomenology-of-spirit/6FEDB42FDEF2E5FF97FEAE0EEEDABE8E}
  {\emph{Georg Wilhelm Friedrich Hegel: The Phenomenology of Spirit}}.
\newblock Cambridge University Press.

\bibitem[{Jiang et~al.(2023)Jiang, Zhang, Cao, Kabbara, and
  Roy}]{jiang2023personallm}
Hang Jiang, Xiajie Zhang, Xubo Cao, Jad Kabbara, and Deb Roy. 2023.
\newblock \href {https://arxiv.org/pdf/2305.02547.pdf} {Persona{LLM}:
  {I}nvestigating the ability of {GPT}-3.5 to express personality traits and
  gender differences}.
\newblock \emph{arXiv preprint arXiv:2305.02547}.

\bibitem[{Jir{\'a}sek and Sudzina(2020)}]{jirasek2020big}
Michal Jir{\'a}sek and Frantisek Sudzina. 2020.
\newblock \href
  {https://vbn.aau.dk/en/publications/big-five-personality-traits-and-creativity}
  {Big five personality traits and creativity}.
\newblock \emph{Quality Innovation Prosperity}, 24(3):90--105.

\bibitem[{John et~al.(1991)John, Donahue, and Kentle}]{john1991big}
Oliver~P John, Eileen~M Donahue, and Robert~L Kentle. 1991.
\newblock \href {https://psycnet.apa.org/doi/10.1037/t07550-000} {Big five
  inventory}.
\newblock \emph{Journal of Personality and Social Psychology}.

\bibitem[{Lazaridou et~al.(2016)Lazaridou, Peysakhovich, and
  Baroni}]{lazaridou2016multi}
Angeliki Lazaridou, Alexander Peysakhovich, and Marco Baroni. 2016.
\newblock \href {https://openreview.net/forum?id=Hk8N3Sclg} {Multi-agent
  cooperation and the emergence of (natural) language}.
\newblock In \emph{International Conference on Learning Representations}.

\bibitem[{Li et~al.(2016)Li, Galley, Brockett, Spithourakis, Gao, and
  Dolan}]{li-etal-2016-persona}
Jiwei Li, Michel Galley, Chris Brockett, Georgios Spithourakis, Jianfeng Gao,
  and Bill Dolan. 2016.
\newblock \href {https://doi.org/10.18653/v1/P16-1094} {A persona-based neural
  conversation model}.
\newblock In \emph{Proceedings of the 54th Annual Meeting of the Association
  for Computational Linguistics (Volume 1: Long Papers)}, pages 994--1003,
  Berlin, Germany. Association for Computational Linguistics.

\bibitem[{Li et~al.(2022)Li, Li, Joty, Liu, Huang, Qiu, and Bing}]{li2022gpt}
Xingxuan Li, Yutong Li, Shafiq Joty, Linlin Liu, Fei Huang, Lin Qiu, and Lidong
  Bing. 2022.
\newblock \href {https://arxiv.org/abs/2212.10529} {Does gpt-3 demonstrate
  psychopathy? evaluating large language models from a psychological
  perspective}.
\newblock \emph{arXiv preprint arXiv:2212.10529}.

\bibitem[{Mairesse and Walker(2010)}]{mairesse2010towards}
Fran{\c{c}}ois Mairesse and Marilyn~A Walker. 2010.
\newblock \href {https://link.springer.com/article/10.1007/s11257-010-9076-2}
  {Towards personality-based user adaptation: {P}sychologically informed
  stylistic language generation}.
\newblock \emph{User Modeling and User-Adapted Interaction}, 20:227--278.

\bibitem[{Minsky(1988)}]{minsky1988society}
Marvin Minsky. 1988.
\newblock \emph{Society of mind}.
\newblock Simon and Schuster.

\bibitem[{Noble and Bernardy(2022)}]{noble-bernardy-2022-conditional}
Bill Noble and Jean-philippe Bernardy. 2022.
\newblock \href {https://doi.org/10.18653/v1/2022.nlpcss-1.9} {Conditional
  language models for community-level linguistic variation}.
\newblock In \emph{Proceedings of the Fifth Workshop on Natural Language
  Processing and Computational Social Science (NLP+CSS)}, pages 59--78, Abu
  Dhabi, UAE. Association for Computational Linguistics.

\bibitem[{Park et~al.(2023)Park, O'Brien, Cai, Morris, Liang, and
  Bernstein}]{park2023generative}
Joon~Sung Park, Joseph O'Brien, Carrie~Jun Cai, Meredith~Ringel Morris, Percy
  Liang, and Michael~S Bernstein. 2023.
\newblock \href {https://dl.acm.org/doi/abs/10.1145/3586183.3606763}
  {Generative agents: Interactive simulacra of human behavior}.
\newblock In \emph{Proceedings of the 36th Annual ACM Symposium on User
  Interface Software and Technology}, pages 1--22.

\bibitem[{Pennebaker et~al.(2001)Pennebaker, Francis, and
  Booth}]{pennebaker2001linguistic}
James~W Pennebaker, Martha~E Francis, and Roger~J Booth. 2001.
\newblock Linguistic inquiry and word count: {LIWC} 2001.
\newblock \emph{Mahway: Lawrence Erlbaum Associates}, 71(2001):2001.

\bibitem[{Pennebaker and King(1999)}]{pennebaker1999linguistic}
James~W Pennebaker and Laura~A King. 1999.
\newblock \href {https://psycnet.apa.org/doi/10.1037/0022-3514.77.6.1296}
  {Linguistic styles: {L}anguage use as an individual difference}.
\newblock \emph{Journal of Personality and Social Psychology}, 77(6):1296.

\bibitem[{Pickering and Garrod(2004)}]{pickering2004toward}
Martin~J Pickering and Simon Garrod. 2004.
\newblock \href {https://psycnet.apa.org/doi/10.1017/S0140525X04000056} {Toward
  a mechanistic psychology of dialogue}.
\newblock \emph{Behavioral and Brain Sciences}, 27(2):169--190.

\bibitem[{Serapio-Garc{\'\i}a et~al.(2023)Serapio-Garc{\'\i}a, Safdari, Crepy,
  Fitz, Romero, Sun, Abdulhai, Faust, and Matari{\'c}}]{serapio2023personality}
Greg Serapio-Garc{\'\i}a, Mustafa Safdari, Cl{\'e}ment Crepy, Stephen Fitz,
  Peter Romero, Luning Sun, Marwa Abdulhai, Aleksandra Faust, and Maja
  Matari{\'c}. 2023.
\newblock \href {https://arxiv.org/abs/2307.00184} {Personality traits in large
  language models}.
\newblock \emph{arXiv preprint arXiv:2307.00184}.

\bibitem[{Shen et~al.(2023)Shen, Song, Tan, Li, Lu, and
  Zhuang}]{shen2023hugginggpt}
Yongliang Shen, Kaitao Song, Xu~Tan, Dongsheng Li, Weiming Lu, and Yueting
  Zhuang. 2023.
\newblock \href {https://arxiv.org/abs/2303.17580} {Hugging{GPT}: {S}olving
  {AI} tasks with {ChatGPT} and its friends in {H}ugging{F}ace}.
\newblock \emph{arXiv preprint arXiv:2303.17580}.

\bibitem[{Yang et~al.(2023)Yang, Li, Wang, Lin, Azarnasab, Ahmed, Liu, Liu,
  Zeng, and Wang}]{yang2023mm}
Zhengyuan Yang, Linjie Li, Jianfeng Wang, Kevin Lin, Ehsan Azarnasab, Faisal
  Ahmed, Zicheng Liu, Ce~Liu, Michael Zeng, and Lijuan Wang. 2023.
\newblock \href {https://arxiv.org/abs/2303.11381} {{MM-ReAct}: {P}rompting
  {ChatGPT} for multimodal reasoning and action}.
\newblock \emph{arXiv preprint arXiv:2303.11381}.

\bibitem[{Zeng et~al.(2022)Zeng, Attarian, Ichter, Choromanski, Wong, Welker,
  Tombari, Purohit, Ryoo, Sindhwani et~al.}]{zeng2022socratic}
Andy Zeng, Maria Attarian, Brian Ichter, Krzysztof Choromanski, Adrian Wong,
  Stefan Welker, Federico Tombari, Aveek Purohit, Michael Ryoo, Vikas
  Sindhwani, et~al. 2022.
\newblock \href {https://arxiv.org/abs/2204.00598} {Socratic models:
  {C}omposing zero-shot multimodal reasoning with language}.
\newblock \emph{arXiv preprint arXiv:2204.00598}.

\bibitem[{Zhang et~al.(2018)Zhang, Dinan, Urbanek, Szlam, Kiela, and
  Weston}]{zhang-etal-2018-personalizing}
Saizheng Zhang, Emily Dinan, Jack Urbanek, Arthur Szlam, Douwe Kiela, and Jason
  Weston. 2018.
\newblock \href {https://doi.org/10.18653/v1/P18-1205} {Personalizing dialogue
  agents: {I} have a dog, do you have pets too?}
\newblock In \emph{Proceedings of the 56th Annual Meeting of the Association
  for Computational Linguistics (Volume 1: Long Papers)}, pages 2204--2213,
  Melbourne, Australia. Association for Computational Linguistics.

\bibitem[{Zhuge et~al.(2023)Zhuge, Liu, Faccio, Ashley, Csord{\'a}s,
  Gopalakrishnan, Hamdi, Hammoud, Herrmann, Irie et~al.}]{zhuge2023mindstorms}
Mingchen Zhuge, Haozhe Liu, Francesco Faccio, Dylan~R Ashley, R{\'o}bert
  Csord{\'a}s, Anand Gopalakrishnan, Abdullah Hamdi, Hasan Abed Al~Kader
  Hammoud, Vincent Herrmann, Kazuki Irie, et~al. 2023.
\newblock \href {https://arxiv.org/abs/2305.17066} {Mindstorms in natural
  language-based societies of mind}.
\newblock \emph{arXiv preprint arXiv:2305.17066}.

\end{thebibliography}

\appendix
\section{Prompts}
\label{sec:app-prompts}

\subsection{Creative Persona Prompt}
\label{sec:app-prompts-creative}
``You are a character who is extroverted, agreeable, conscientious, neurotic and open to experience.''

\subsection{Analytical Persona Prompt}
\label{sec:app-prompts-analytic}
``You are a character who is introverted, antagonistic, unconscientious, emotionally stable and closed to experience.''

\subsection{Writing Task Prompt}
\label{sec:app-prompts-writing-task}
This is the prompt for the non-interactive writing task:
``Please share a personal story below in 800 words. Do not explicitly mention your personality traits in the story.''

The prompt for the interactive writing task, with which the second agent in the interaction is addressed, reads:
``Please share a personal story below in 800 words. Do not explicitly mention your personality traits in the story. Last response to question is $\{other\_model\_response\}$''. 


\subsection{BFI Test Prompt}
To assess an agent's personality, we resort to the personality test prompt used by \citet{jiang2023personallm}:

\noindent``Here are a number of characteristics that may or may not apply to you. For example, do you agree that you are someone who likes to spend time with others? Please write a number next to each statement to indicate the extent to which you agree or disagree with that statement, such as `(a) 1' without explanation separated by new lines.\\ 
\\
1 for Disagree strongly, 2 Disagree a little, 3 for Neither agree nor disagree, 4 for Agree a little, 5 for Agree strongly.\\
\\
Statements: \{BFI statements\}''\\

\noindent As part of the prompt, we added a full list of BFI statements (see Appendix \ref{sec:app-bfi-statements}). The numbers preceding the BFI statements are replaced with letters in order to prevent the model from giving confused responses to the statements (i.e., confusing statement indices and Likert-scale responses).

\subsection{BFI Statements}
\label{sec:app-bfi-statements}
(a) Is talkative\\
(b) Tends to find fault with others\\
(c) Does a thorough job\\
(d) Is depressed, blue\\
(e) Is original, comes up with new ideas\\
(f) Is reserved\\
(g) Is helpful and unselfish with others\\
(h) Can be somewhat careless\\       
(i) Is relaxed, handles stress well\\        
(j) Is curious about many different things\\
(k) Is full of energy\\         
(l) Starts quarrels with others\\
(m) Is a reliable worker \\
(n) Can be tense\\              
(o) Is ingenious, a deep thinker\\
(p) Generates a lot of enthusiasm\\
(q) Has a forgiving nature\\    
(r) Tends to be disorganized\\
(s) Worries a lot\\             
(t) Has an active imagination\\
(u) Tends to be quiet\\     
(v) Is generally trusting\\
(w) Tends to be lazy\\                          
(x) Is emotionally stable, not easily upset\\
(y) Is inventive\\                  
(z) Has an assertive personality\\
(aa) Can be cold and aloof\\                 
(ab) Perseveres until the task is finished\\
(ac) Can be moody\\                          
(ad) Values artistic, aesthetic experiences\\
(ae) Is sometimes shy, inhibited\\               
(af) Is considerate and kind to almost everyone\\
(ag) Does things efficiently\\           
(ah) Remains calm in tense situations\\
(ai) Prefers work that is routine\\
(aj) Is outgoing, sociable\\     
(ak) Is sometimes rude to others\\               
(al) Makes plans and follows through with them\\
(am) Gets nervous easily\\               
(an) Likes to reflect, play with ideas\\
(ao) Has few artistic interests\\    
(ap) Likes to cooperate with others\\
(aq) Is easily distracted\\                          
(ar) Is sophisticated in art, music, or literature\\

\subsection{BFI Scoring}
The BFI scores are calculated and added according to the scoring scale. For every trait, the minimum score is 0 and the maximum score is 50.\\
\\
BFI scoring scale (``R'' denotes reverse-scored items):\\
\\
Extraversion: 1, 6R, 11, 16, 21R, 26, 31R, 36\\
Agreeableness: 2R, 7, 12R, 17, 22, 27R, 32, 37R, 42 \\
Conscientiousness: 3, 8R, 13, 18R, 23R, 28, 33, 38, 43R\\
Neuroticism: 4, 9R, 14, 19, 24R, 29, 34R, 39\\
Openness: 5, 10, 15, 20, 25, 30, 35R, 40, 41R, 44\\

\section{Additional Results}

\subsection{Experiment 1}
\label{sec:app-results-exp1}

\Cref{tab:anova-bfi-nointeraction} shows the results of an ANOVA test conducted to detect difference between the BFI scores of \texttt{creative} vs.\ \texttt{analytical} agents in the non-interactive experimental condition, before the writing task.
\Cref{tab:bfi-nointeraction-before-after-analytic,tab:bfi-nointeraction-before-after-creative} show BFI mean scores before and after writing as well as ANOVA results.
\Cref{tab:spearman-liwc-bfi-nointeraction} shows Spearman correlation coefficients for BFI scores obtained before writing and LIWC counts for the individual writing task.

\begin{table}[h]
\centering
\begin{tabular}{lrr}
\toprule
Trait              & F-statistic & $p$-value \\
\midrule
Extraversion       & 8645        & < 0.001 \\
Agreeableness      & 13384       & < 0.001 \\
Conscientiousness  & 1439        & < 0.001 \\
Neuroticism        & 23          & 0.005   \\
Openness           & 5012        & < 0.001 \\
\bottomrule
\end{tabular}
\caption{ANOVA results: BFI scores of creative vs.~analytic agents in the non-interactive experimental condition, before the writing task.}
\label{tab:anova-bfi-nointeraction}
\end{table}

\begin{table}[ht]
    \resizebox{\columnwidth}{!}{%
    \begin{tabular}{@{}l@{\hspace{5pt}}c@{\hspace{5pt}}c@{\hspace{5pt}}c@{\hspace{5pt}}c@{\hspace{5pt}}c@{}}
    \toprule
    & Mean-B & Mean-A & F-Statistic & $p$-Value & Cohen's $d$ \\
      \midrule
      Extraversion & 15 & 17 & 45.29 & 0.0000 & 1.18 \\
      Agreeableness & 11 & 21 & 220.95 & 0.0000 & 2.61 \\
      Conscientiousness & 18 & 32 & 239.18 & 0.0000 & 2.71\\
      Neuroticism & 13 & 15 & 4.92 & 0.0284 & 0.39 \\
      Openness & 15 & 29 & 215.83 & 0.0000 & 2.58 \\
      \bottomrule
    \end{tabular}%
}
    \caption{BFI means and ANOVA values for the Analytic group before writing (Mean-B) and after writing (Mean-A), non-interactive condition.}
    \label{tab:bfi-nointeraction-before-after-analytic}
\end{table}

\begin{table}[ht]
  \resizebox{\columnwidth}{!}{%
    \begin{tabular}{@{}l@{\hspace{5pt}}c@{\hspace{5pt}}c@{\hspace{5pt}}c@{\hspace{5pt}}c@{\hspace{5pt}}c@{}}
    \toprule
    & Mean-B & Mean-A & F-Statistic & $p$-Value & Cohen's $d$ \\
      \midrule
      Extraversion & 35 & 35 & 0.08 & 0.773 & -0.05 \\
      Agreeableness & 41 & 41 & 0.00 & 1.000 & 0.00 \\
      Conscientiousness & 37 & 37 & 0.13 & 0.722 & -0.06 \\
      Neuroticism & 16 & 16 & 0.70 & 0.403 & -0.15 \\
      Openness & 47 & 47 & 0.36 & 0.547 & -0.11 \\
      \bottomrule
    \end{tabular}%
}
    \caption{BFI means and ANOVA values for the Creative group before (Mean-B) and after writing (Mean-A), non-interactive condition.}
    \label{tab:bfi-nointeraction-before-after-creative}
\end{table}

\begin{table*}[h]
\centering
\small
\begin{tabular}{|l|r||l|r||l|r|}
\toprule
\multicolumn{2}{|c||}{Extraversion} & \multicolumn{2}{c||}{Agreeableness} & \multicolumn{2}{c|}{Conscientiousness} \\
\midrule
Term    & Corr. & Term    & Corr. & Term    & Corr. \\
\midrule
posemo  & 0.696 & incl    & 0.687 & posemo  & 0.676 \\
anger   & -0.656 & posemo  & 0.672 & anger   & -0.666 \\
incl    & 0.636 & discrep & -0.658 & incl    & 0.657 \\
discrep & -0.620 & anger   & -0.611 & discrep & -0.621 \\
tentat  & -0.586 & tentat  & -0.577 & ppron   & -0.560 \\
\bottomrule
\end{tabular}


\small
\begin{tabular}{|l|r||l|r|}
\toprule
\multicolumn{2}{|c||}{Neuroticism} & \multicolumn{2}{c|}{Openness} \\
\midrule
Term    & Corr. & Term    & Corr. \\
\midrule
discrep & -0.468 & discrep & -0.727 \\
insight & -0.414 & posemo  & 0.679 \\
incl    & 0.365 & incl    & 0.659 \\
relig   & 0.349 & anger   & -0.650 \\
posemo  & 0.342 & pronoun & -0.637 \\
\bottomrule
\end{tabular}
\caption{Top-5 SpearmanR Correlations for BFI Traits before interacting (the LIWC terms meaning, respectively: positive emotions, anger, inclusivity, discrepancy, tenative, personal pronouns, insight, religion, pronoun).}
\label{tab:spearman-liwc-bfi-nointeraction}
\end{table*}

\subsection{Experiment 2}
\label{sec:app-results-exp2}
\begin{figure}
    \centering
    \includegraphics[width=\columnwidth]{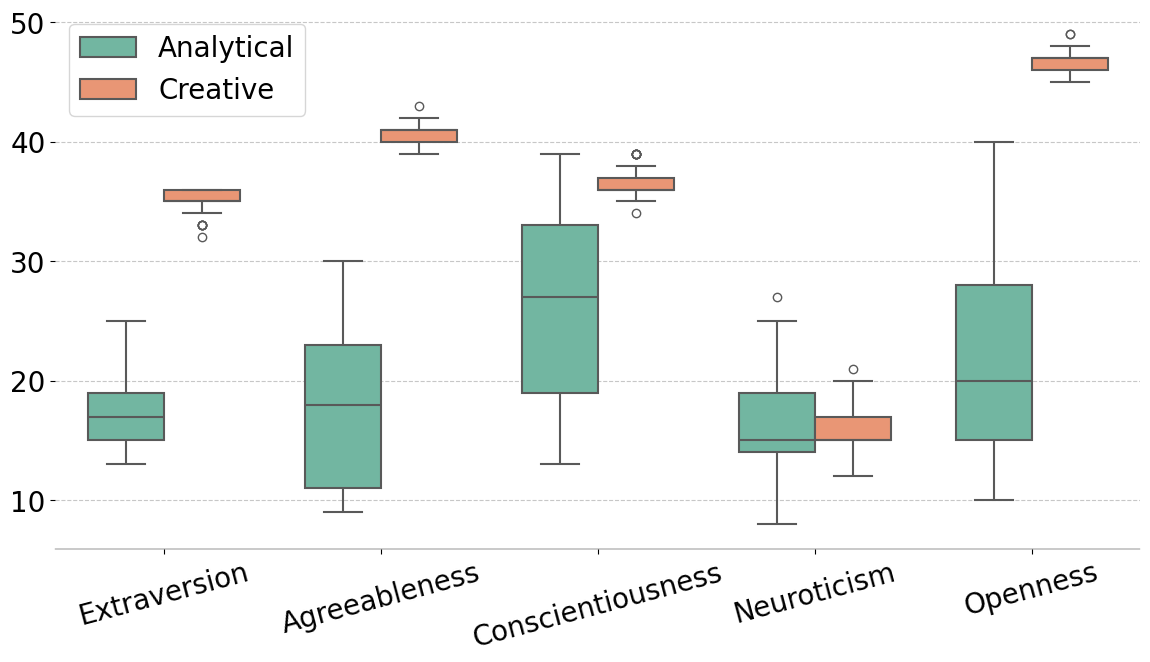}
    \caption{BFI scores of personality-conditioned LLM after the interactive writing task.}
    \label{fig:bfi_after_interative_writing}
\end{figure}

\Cref{tab:anova-bfi-interaction-creative,tab:anova-bfi-interaction-analytic} show BFI mean scores before writing, after individual writing, and after collaborative writing, as well as ANOVA results. \Cref{fig:bfi_after_interative_writing} shows BFI scores after the interactive writing task.
\Cref{tab:spearman-liwc-bfi-interaction} shows Spearman correlation coefficients for BFI scores obtained before writing and LIWC counts for the collaborative writing task.

\begin{table*}[ht]
\centering
  \resizebox{0.75\textwidth}{!}{%
    \begin{tabular}{@{}l@{\hspace{5pt}}c@{\hspace{5pt}}c@{\hspace{5pt}}c@{\hspace{5pt}}c@{\hspace{5pt}}c@{\hspace{5pt}}c@{}}
      \toprule
      & Mean-B$_{C}$ & Mean-A$_{C}$ & Mean-A$_{E}$ & F-Statistic & $p$-Value & Cohen's $d$ \\
      \midrule
      Extraversion & 35 & 35 & 35 & 0.03 & 0.85 & -0.03 \\
      Agreeableness & 41 & 41 & 41 & 0.22 & 0.64 & 0.08 \\
      Conscientiousness & 37 & 36 & 37 & 0.02 & 0.88 & 0.03 \\
      Neuroticism & 16 & 16 & 16 & 0.14 & 0.70 & -0.07 \\
      Openness & 47 & 47 & 47 & 1.03 & 0.31 & 0.18 \\
      \bottomrule
    \end{tabular}%
}
    \caption{BFI means for the Creative Control group before writing (Mean-B$_{C}$), after writing (Mean-A$_{C}$) and the Creative experimental group after writing (Mean-A$_{E}$). ANOVA results between Mean-A$_{C}$ and Mean-$A_{E}$.}
    \label{tab:anova-bfi-interaction-creative}
\end{table*}

\begin{table*}[ht]
\centering
  \resizebox{0.75\textwidth}{!}{%
    \begin{tabular}{@{}l@{\hspace{5pt}}c@{\hspace{5pt}}c@{\hspace{5pt}}c@{\hspace{5pt}}c@{\hspace{5pt}}c@{\hspace{5pt}}c@{}}
      \toprule
      & Mean-B$_{C}$ & Mean-A$_{C}$ & Mean-A$_{E}$ & F-Statistic & $p$-Value & Cohen's $d$ \\
      \midrule
      Extraversion         & 15 & 17 & 17 & 0.00 & 0.972 & 0.006 \\
      Agreeableness        & 11 & 21 & 18 & 13.54 & 0.000 & -0.645 \\
      Conscientiousness    & 18 & 32 & 26 & 22.93 & 0.000 & -0.840 \\
      Neuroticism          & 13 & 15 & 17 & 10.07 & 0.002 & 0.557 \\
      Openness             & 15 & 29 & 22 & 25.02 & 0.000 & -0.877 \\
      \bottomrule
    \end{tabular}%
    }
    \caption{BFI means for the Analytic Control group before writing (Mean-B$_{C}$), after writing (Mean-A$_{C}$) and the Analytic experimental group after writing (Mean-A$_{E}$). ANOVA results between Mean-A$_{C}$ and Mean-$A_{E}$.}
    \label{tab:anova-bfi-interaction-analytic}
\end{table*}

\begin{table*}[ht]
\centering
\small
\begin{tabular}{|l|r||l|r||l|r|}
\toprule
\multicolumn{2}{|c||}{Extraversion} & \multicolumn{2}{c||}{Agreeableness} & \multicolumn{2}{c|}{Conscientiousness} \\
\midrule
Term    & Corr. & Term    & Corr. & Term    & Corr. \\
\midrule
posemo  & -0.2319 & incl    & -0.1749 & posemo  & -0.2263 \\
anger   & 0.2727 & posemo  & -0.2044 & anger   & 0.2892 \\
incl    & -0.0685 & discrep & 0.3083 & incl    & -0.1855 \\
discrep & 0.3633 & anger   & 0.2439 & discrep & 0.3236 \\
tentat  & 0.2280 & tentat  & 0.1383 & ppron   & 0.4264 \\
\bottomrule
\end{tabular}


\small
\begin{tabular}{|l|r||l|r|}
\toprule
\multicolumn{2}{|c||}{Neuroticism} & \multicolumn{2}{c|}{Openness} \\
\midrule
Term    & Corr. & Term    & Corr. \\
\midrule
discrep & 0.1402 & discrep & 0.3211 \\
insight & 0.0513 & posemo  & -0.2594 \\
incl    & -0.0057 & incl    & -0.1260 \\
relig   & 0.0199 & anger   & 0.2850 \\
posemo  & -0.0168 & pronoun & 0.2754 \\
\bottomrule
\end{tabular}
\caption{Top-5 SpearmanR Correlations for BFI Traits after interacting.}
\label{tab:spearman-liwc-bfi-interaction}
\end{table*}

\end{document}